\documentclass[runningheads]{llncs}
\usepackage[T1]{fontenc}
\usepackage{graphicx}
\usepackage{color}
\usepackage[colorlinks=true, 
linkcolor=blue, 
citecolor=blue, 
urlcolor=blue]{hyperref}

\usepackage{amsmath,amssymb,amsfonts}
\usepackage{algorithmic,multirow,graphicx,textcomp,tikz,booktabs,subcaption,overpic,bm}
\begin{document}
\title{Normalisation and Initialisation Strategies for Graph Neural Networks in Blockchain Anomaly Detection}
\titlerunning{Graph Neural Networks in Blockchain Anomaly Detection}
%
\author{Dang Sy Duy\inst{1} \and
Nguyen Duy Chien\inst{2} \and
Kapil Dev\inst{1} \and
Jeff Nijsse\inst{1}}
\authorrunning{D. Dang et al.}
\institute{RMIT University, Hanoi, Vietnam\\
\email{dduy193.cs@gmail.com, \{kapil.dev, jeff.nijsse\}@rmit.edu.vn} \and
Vietnam National University, Hanoi, Vietnam\\
\email{duychien.research@gmail.com}}
\maketitle              
\begin{abstract}
Graph neural networks (GNNs) offer a principled approach to financial fraud detection by jointly learning from node features and transaction graph topology. However, their effectiveness on real-world anti-money laundering (AML) benchmarks depends critically on training practices such as specifically weight initialisation and normalisation that remain underexplored. We present a systematic ablation of initialisation and normalisation strategies across three GNN architectures (GCN, GAT, and GraphSAGE) on the Elliptic Bitcoin dataset. Our experiments reveal that initialisation and normalisation are architecture-dependent: GraphSAGE achieves the strongest performance with Xavier initialisation alone, GAT benefits most from combining GraphNorm with Xavier initialisation, while GCN shows limited sensitivity to these modifications. These findings offer practical, architecture-specific guidance for deploying GNNs in AML pipelines for datasets with severe class imbalance. We release a reproducible experimental framework with temporal data splits, seeded runs, and full ablation results.

\end{abstract}
%
%
%
\section{Introduction}\label{sec:introduction}
Financial transaction data naturally forms graphs, where illicit behaviour often emerges from coordinated structures rather than isolated records. On datasets such as Elliptic \cite{weber2019amlgnn}, traditional i.i.d.\ models (e.g., logistic regression, random forests) overlook these relational patterns, whereas graph neural networks (GNNs) can leverage both node features and graph topology \cite{hamilton2017inductive,kipf2016semi}. However, the effectiveness of GNNs in this context depends critically on two often-overlooked factors: \emph{initialisation} and \emph{normalisation}.

Despite the promise of GNNs for fraud detection, several open questions remain regarding their effectiveness and the best practices for their deployment in real-world transaction graphs. To address these gaps, we focus on the following research questions:

\begin{enumerate}
    \item What properties of transaction graphs (e.g., heterophily, degree skew, temporal drift) limit GNN effectiveness on Elliptic-like data?
    \item How do initialisation strategies affect optimisation stability and convergence?
    \item Can graph-specific normalisation reduce feature drift and over-smoothing more effectively than batch or layer normalisation?
    \item Which normalisation methods are most effective across GCN, Graph Attention Network (GAT), and GraphSAGE for fraud detection?
\end{enumerate}

We approach these questions by analysing how standard initialisations (Xavier \cite{glorot2010understanding}, Kaiming \cite{he2015delving}) interact with repeated neighbourhood aggregation, and by evaluating normalisation schemes tailored to graphs. Unlike batch or layer normalisation which assume independent samples, graph-aware methods such as GraphNorm \cite{cai2020graphnorm} normalise in the context of the graph, aiming to preserve variance, mitigate over-smoothing, and curb high-degree dominance. Experiments use temporally correct splits to reflect realistic deployment and avoid leakage.

The investigation leads to the following contributions:

\noindent
\begin{enumerate}
    \item \textbf{Systematic analysis of initialisation and graph-level normalisation in AML graphs.} We conduct a controlled ablation of Xavier initialisation and GraphNorm across GCN, GAT, and GraphSAGE on the temporally structured and highly imbalanced Elliptic dataset.
    \item \textbf{Architecture-dependent insights on training dynamics.} Our results show that the effects of initialisation and normalisation are not uniform: GraphSAGE achieves the strongest performance with Xavier initialisation, GAT benefits from GraphNorm + Xavier, while GCN shows limited gains from additional graph-level normalisation.
    \item \textbf{Reproducible framework.} We adopt leakage-free temporal splits, AUPRC-focused evaluation, and repeated subsampling to provide robust and deployment-relevant assessment for AML scenarios.
\end{enumerate}

\section{Background}\label{sec:background}

\subsection{Graph Neural Networks for Financial Fraud Detection}
GNNs have become an important tool in financial fraud detection because they use both node features and the relationships in transaction networks. Weber et al.~\cite{weber2019amlgnn} first applied Graph Convolutional Networks (GCNs) to Bitcoin transactions and showed improvements over traditional machine learning methods based on handcrafted features. Dwivedi et al.~\cite{dwivedi2020benchmarkgnns} compared several GNNs for fraud detection and found that architectures such as GAT~\cite{velickovic2018graph} and GraphSAGE~\cite{hamilton2017inductive} respond differently to graph structure, making model choice an important factor. Later work introduced temporal models, such as Temporal Graph Networks (TGNs)~\cite{rossi2020temporal}, which use time information to improve detection accuracy compared with static models, though at the cost of higher computation. Scalability has also been addressed by sampling-based methods such as GraphSAINT~\cite{zeng2019graphsaint} and FastGCN~\cite{chen2018fastgcn}, which allow training on large graphs but may reduce accuracy under strong class imbalance.  

Much of the prior work has focused on architectural design or scalability, while training practices such as weight initialisation and normalisation remain less explored. Approaches like GraphNorm~\cite{cai2020graphnorm} and initialisation schemes such as Xavier~\cite{glorot2010understanding} and He~\cite{he2015delving} have shown benefits in other domains, such as text relevance estimation \cite{selim2024}, and spatial-temporal analysis \cite{noval2024}, but their role in fraud detection is still unclear. This gap is important because class imbalance and node dependencies in transaction graphs can strongly influence how models behave under different training conditions. Our work addresses this by analysing how initialisation and normalisation affect GNN training and performance in financial fraud detection.  

We analyse three widely used GNN architectures: GCN, GAT, and GraphSAGE. GCN, introduced by Kipf and Welling \cite{kipf2016semi}, aggregates neighbour features using normalised adjacency matrices and is valued for its efficiency, making it a common baseline in fraud detection \cite{weber2019amlgnn,alarab2020}. GATs \cite{velickovic2018graph} extend this by assigning different weights to neighbors with an attention mechanism, allowing the model to focus on more relevant transactions or accounts. GraphSAGE \cite{hamilton2017inductive} provides an inductive framework that generates embeddings for unseen nodes by sampling and aggregating neighbors with operators such as mean, max-pooling, or LSTM-based aggregators. This property makes it suitable for evolving financial networks \cite{marasi2024}.

\subsection{Normalisation Techniques}
Normalisation helps stabilise neural network training and improve convergence. BatchNorm~\cite{ioffe2015batch} normalises activations in each mini-batch using the mean and variance, but it assumes samples are independent, which is not the case in graph data. LayerNorm~\cite{ba2016layer} normalises features within each sample, removing dependence on batch size but still ignoring graph structure.  

Graph-specific methods address this limitation. GraphNorm~\cite{cai2020graphnorm} uses graph-level statistics to normalise features, helping preserve variance and reduce over-smoothing, where node representations become too similar in deeper networks. Other methods such as PairNorm~\cite{zhao2020pairnorm} and Differentiable Group Normalisation (DGN)~\cite{zhou2020dgn} also aim to maintain distinctions between nodes.

\subsection{Weight Initialisation in Graph Neural Networks}
Weight initialisation affects how quickly and reliably neural networks train. Xavier initialisation~\cite{glorot2010understanding} maintains the variance of activations across layers and reduces the risk of vanishing or exploding gradients. He initialisation~\cite{he2015delving}, designed for ReLU-based networks, increases the initial variance to account for the properties of the ReLU function, improving stability in deeper models.  

In GNNs, Xavier initialisation is the most common. It was used in the original GCN by Kipf and Welling~\cite{kipf2016semi} and in GraphSAGE by Hamilton et al.~\cite{hamilton2017inductive}, supporting effective feature propagation and inductive learning. Despite this, systematic studies of initialisation in GNNs are limited. Poor initialisation can hinder gradient flow, slow convergence, and reduce accuracy, making it important to evaluate initialisation strategies for financial fraud detection tasks.  

\section{Methodology}\label{sec:methodology}
\subsection{The Elliptic Dataset Benchmark}
The Elliptic Bitcoin Dataset~\cite{weber2019amlgnn} is a widely used benchmark for AML tasks in cryptocurrency networks such as evaluating GCNs \cite{weber2019amlgnn}, architectural benchmarks \cite{dwivedi2020benchmarkgnns}, temporal models \cite{rossi2020temporal}, and sampling-based training methods \cite{zeng2019graphsaint,chen2018fastgcn}. It contains 203,769 nodes representing transactions and 234,355 directed edges representing payment flows. Each transaction is described by a 166-dimensional feature vector, including raw features and one-hop aggregated statistics. The dataset is divided into 49 time steps, each covering about two weeks of activity, with edges only within the same time step to allow temporal modeling without information leakage.  

Labels are available for only a subset of nodes: 2\% are illicit, 21\% are licit, and 77\% are unknown. This reflects the strong imbalance seen in real-world financial crime detection.  

\subsection{Graph Construction}
The directed transaction network is converted to an undirected graph for symmetric message passing. Edges are symmetrised by adding reversed edges, with self-loops and duplicates removed. Transaction IDs are mapped to contiguous integers, and a feature matrix $\mathbf{X} \in \mathbb{R}^{N \times 166}$ is constructed from transaction features. Labels ($0$=illicit, $1$=licit, $2$=unknown) are used for supervised training, with unknown nodes retained for contextual message passing.

\subsection{Temporal Data Splitting}
We apply strict temporal splits to prevent information leakage: 29 time steps for training, 10 for validation, and 10 for testing. Unlike random splits, this ordering ensures predictions are evaluated only on future transactions. Only labeled nodes are used for supervised training, while unlabeled nodes remain in the graph for message passing. We use AUPRC as the primary metric due to severe class imbalance~\cite{hamilton2020,weber2019amlgnn}.

\subsection{Model Architecture}
\begin{figure}[htbp]
\centerline{\includegraphics[width=1\textwidth]{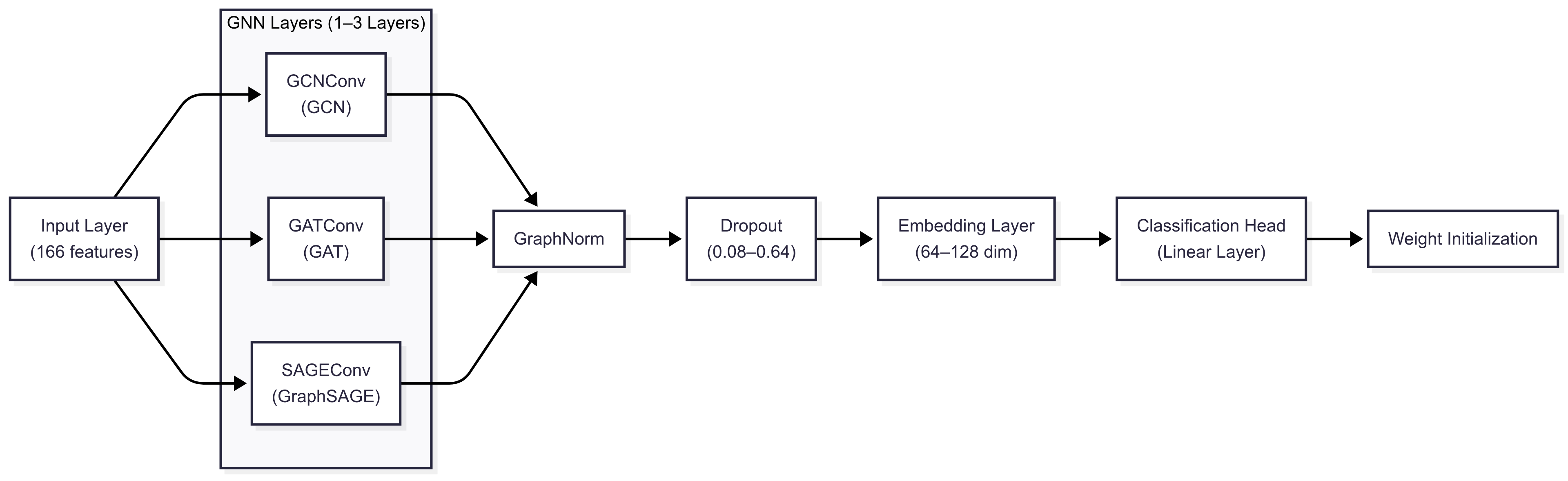}}
\caption{The base model architecture. The evaluated architecture follows a modular design consisting of an input layer (166 transaction features), followed by 1--3 GNN layers with 3 different types (GCN, GAT, and GraphSAGE). Each GNN variant is followed by GraphNorm and dropout (0.08--0.64) for regularisation. The processed features are passed to an embedding layer (64-128 dimensions) and then to a classification head (linear layer) for binary prediction. All layers use standard weight initialisation methods.}
\label{fig:modelarch}
\end{figure}

In the development of machine learning models for AML using the Elliptic Bitcoin Dataset, we evaluate three GNN architectures to assess their effectiveness in classifying Bitcoin transactions as licit or illicit, as shown in Fig.~\ref{fig:modelarch}. The task is formulated as a binary node classification problem, focusing on labeled transactions to identify illicit activities indicative of anomalous behaviours. The three architectures evaluated, GCN, GAT, and GraphSAGE, are selected based on their widespread use in node classification tasks.

Each GNN architecture is designed with the following components. The input layer accepts 166 transaction features, excluding the time step feature, as the model input is defined as $\mathbf{x} = \texttt{features}[:, 1:]$ with shape $[N, 166]$, where $N$ is the number of nodes ($203{,}769$). GNN layers consist of 1 to 3 layers, with hidden dimensions tuned in the range $[128, 256]$ and embedding dimensions in the range $[64, 128]$. These layers perform message passing to aggregate information from neighboring nodes, tailored to each architecture's mechanism. The Classification Head outputs a binary classification (licit or illicit), corresponding to an output dimension of 2, suitable for the supervised learning task on labeled transactions.

The architectures are implemented using PyTorch Geometric, which ensures efficient handling of graph-structured data~\cite{fey2019}. We use specific GNN layers such as \texttt{GCNConv}, \texttt{GATConv}, and \texttt{SAGEConv}, enabling rapid experimentation and flexible model design.







\subsection{Regularisation and Weight Initialisation}
To address overfitting from label imbalance, we apply dropout ($p \in [0.08, 0.64]$) between layers. For normalisation, we evaluate GraphNorm, which computes statistics at the graph level rather than the batch level, reducing sensitivity to structural variations between transaction subgraphs as opposed to standard BatchNorm used in the baseline configuration.

\begin{figure}[htbp]
    \centering
    \begin{subfigure}[t]{0.48\textwidth}
        \centering
        \includegraphics[width=\textwidth]{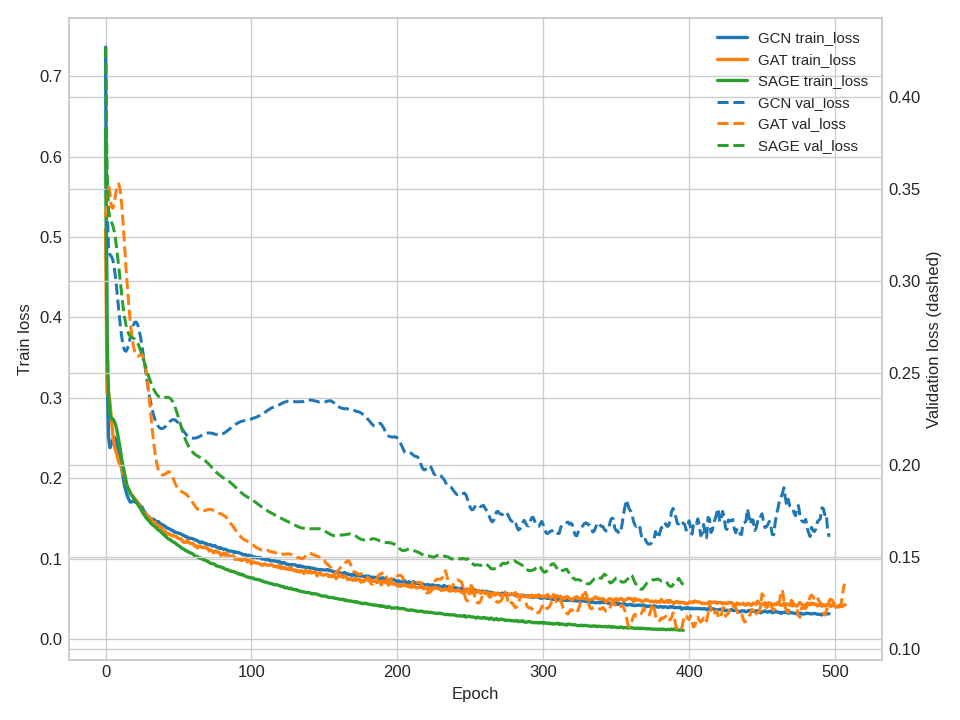}
        \caption{Training loss curves}
        \label{fig:training_loss}
    \end{subfigure}
    \hfill
    \begin{subfigure}[t]{0.48\textwidth}
        \centering
        \includegraphics[width=\textwidth]{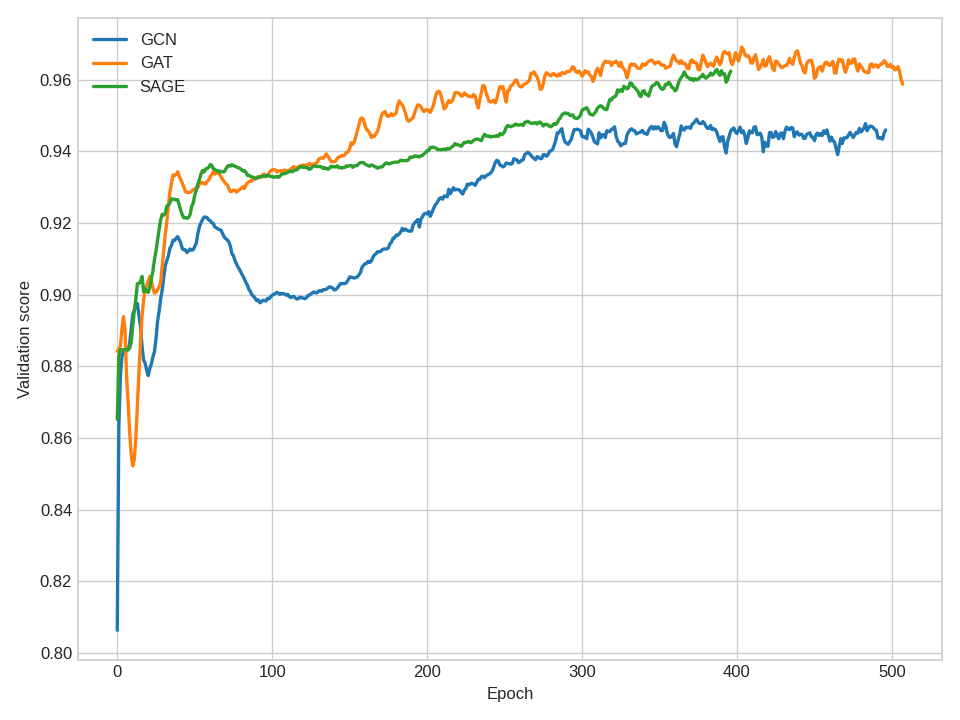}
        \caption{Validation score curves}
        \label{fig:validation_score}
    \end{subfigure}
    \caption{Convergence behaviour of GCN, GAT, and GraphSAGE on the Elliptic dataset. 
    (\subref{fig:training_loss}) Training loss curves show that GraphSAGE converges faster than GCN and GAT. 
    (\subref{fig:validation_score}) Validation score trajectories highlight that GraphSAGE consistently achieves higher validation scores, while GAT exhibits more variance and GCN stabilises more slowly.}
    \label{fig:convergence}
\end{figure}

For weight initialisation, we use Xavier uniform~\cite{glorot2010understanding} for classifier heads and each architecture's default \texttt{reset\_parameters()} for GNN layers. This maintains balanced variance propagation, preventing vanishing or exploding gradients during multi-hop message passing. The convergence behaviour is shown in Fig.~\ref{fig:convergence}.

\subsection{Hyperparameter Optimisation}\label{sec:hparam}
We use \texttt{Optuna} for automated hyperparameter optimisation over 100 trials (Table~\ref{tab:hparam}), employing the Tree-structured Parzen Estimator (TPE) and the Asynchronous Successive Halving Algorithm (ASHA) \cite{akiba2019optuna} to efficiently explore the search space. The optimisation objective is the AUPRC on the validation set, which is more appropriate for model selection under severe class imbalance. The test set is strictly held out and used only for final evaluation after hyperparameter tuning. This protocol prevents information leakage and ensures that reported results reflect generalisation to future, unseen transactions in the temporal setting.

\begin{table}[htbp]
\centering
\caption{Hyperparameter search space and optimal values obtained for each model via Optuna optimisation. Weight decay is fixed at $5 \times 10^{-4}$ for all models.}
\label{tab:hparam}
\begin{tabular}{@{}lccccc@{}}
\toprule
\textbf{Parameter} & \textbf{Range} & \textbf{Scale} & \textbf{GAT} & \textbf{GCN} & \textbf{GraphSAGE} \\
\midrule
Learning rate & $[2 \times 10^{-6}, 10^{-3}]$ & Log & $6.999 \times 10^{-4}$ & $8.475 \times 10^{-4}$ & $5.302 \times 10^{-4}$ \\
Hidden dimension & $[128, 256]$ & Linear & 148 & 211 & 140 \\
Embedding dim. & $[64, 128]$ & Linear & 89 & 90 & 103 \\
Number of layers & $\{1, 2, 3\}$ & Categorical & 2 & 2 & 2 \\
Dropout rate & $[0.08, 0.64]$ & Log & 0.2522 & 0.2361 & 0.1135 \\
Number of epochs & $[128, 512]$ & Linear & 508 & 497 & 397 \\
Aggregator$^{\mathrm{*}}$ & \{mean, max\} & Categorical & -- & -- & mean \\
\bottomrule
\end{tabular}
\end{table}

Continuous parameters such as learning rate are sampled on a logarithmic scale. Small changes in these values can have multiplicative effects on performance, so log-scale sampling gives finer resolution in low-value regions. Dropout rate is also sampled on a logarithmic scale. Lower values may suffice for small GNNs, while deeper architectures require stronger regularisation.

Hidden and embedding dimensions are sampled linearly. Model capacity often grows steadily with dimension size, so linear scaling is appropriate. The number of epochs is also tuned linearly to examine convergence in both short and extended schedules. This matters for imbalanced datasets like Elliptic, where overfitting is more likely due to the scarcity of illicit labels.   

The number of layers and the choice of aggregator are sampled as categorical parameters. The aggregator applies only to GraphSAGE models, enabling architecture-specific tuning.


\subsection{Training Procedure}
The training process follows a full-graph learning paradigm, where the entire transaction graph is processed in each forward pass to preserve multi-hop structural and temporal dependencies critical for fraud detection. This avoids information loss from sampling-based training and retains complete neighbourhood context.

We employ a unified training pipeline across all backbone architectures (GCN, GAT, and GraphSAGE), with architecture-specific hyperparameters selected via Optuna (Section~\ref{sec:hparam}). Each model consists of 1--3 GNN message-passing layers followed by ReLU activation, GraphNorm, and dropout, and a final linear classification head for binary node prediction.

GraphNorm is applied after each message-passing layer to stabilise feature distributions and mitigate over-smoothing in the imbalanced Elliptic transaction graph. Models are trained using Adam, with learning rate, hidden dimension, dropout, and number of epochs optimised through Optuna. Class imbalance in the Elliptic dataset is handled with weighted binary cross-entropy to penalise false negatives more heavily.


\subsection{Implementation Details}
All models are implemented with PyTorch and PyTorch Geometric for efficient processing of graph-structured data. Hyperparameters are tuned with \texttt{Optuna}. Training is performed on a single NVIDIA RTX 5880 CUDA-enabled GPU. Full-graph training is adopted to retain complete neighbourhood information and capture long-range dependencies. This is especially important for the Elliptic dataset, where temporal and transactional relationships often span multiple hops. A global random seed of 42 is fixed for reproducibility. All scripts, configurations, and logs are available on our GitHub.\footnote[1]{Our code is here: \href{https://github.com/RMIT-BDSL/Blockchain-Anomaly-Detection}{https://github.com/RMIT-BDSL/Blockchain-Anomaly-Detection}.}

\subsection{Evaluation}
Reliable evaluation is important in fraud detection, where extreme class imbalance can distort results. We use the AUPRC as the primary metric, since it reflects the model's ability to detect illicit transactions in heavily imbalanced data. Accuracy and AUC-ROC can be misleading in this setting, but we also report AUC-ROC for comparison with prior work.  

To measure robustness, we apply a repeated subsampling strategy. We randomly select 50\% of the test nodes 100 times, compute AUPRC and AUC-ROC for each run, and report the mean and standard deviation. This bootstrapping captures performance variability and provides more reliable estimates of generalisation in settings where transaction distributions may shift over time.

\begin{figure}[htbp]
    \centering
    \begin{subfigure}[t]{0.48\textwidth}
        \centering
        \includegraphics[width=\textwidth]{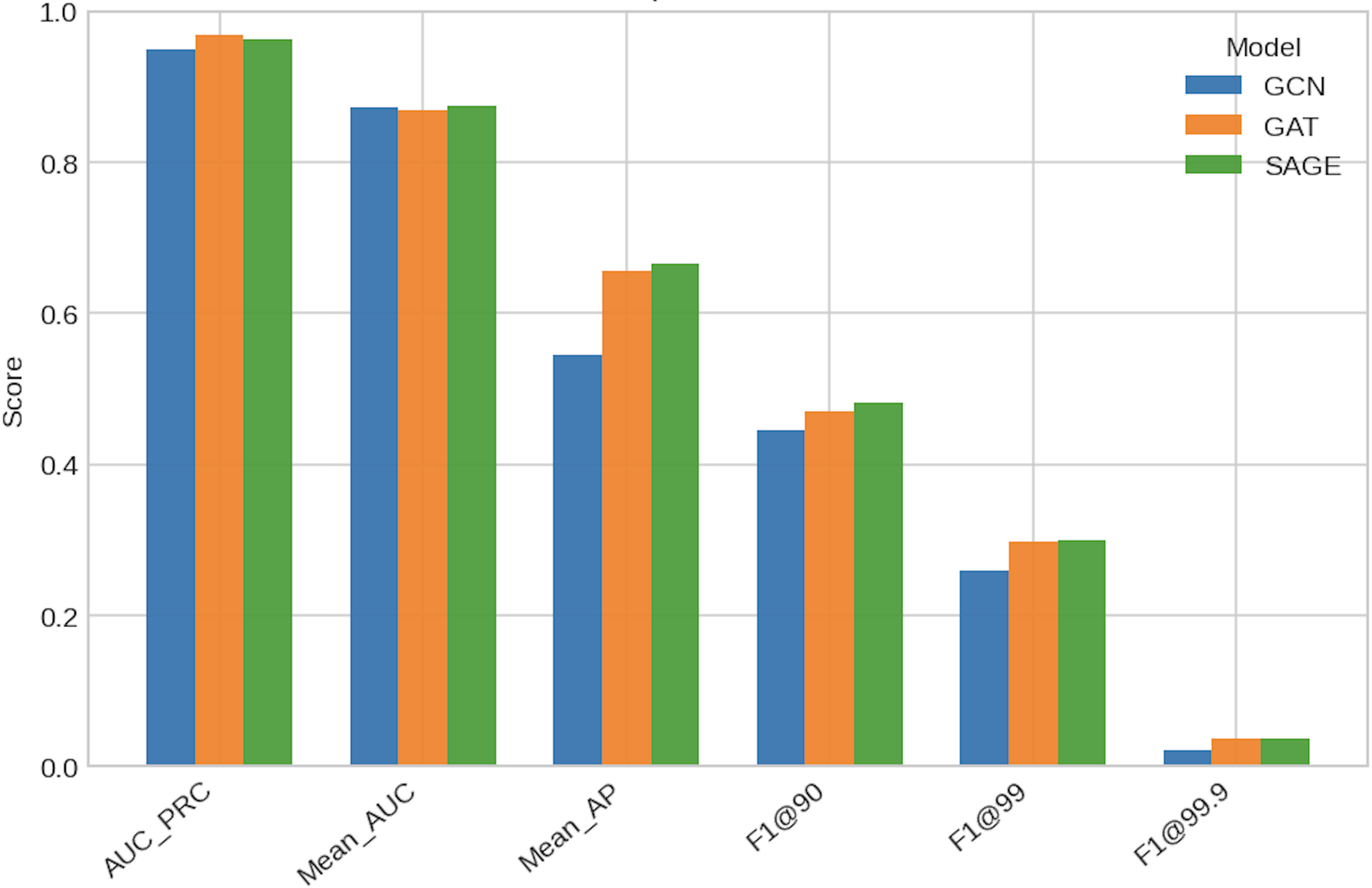}
        \caption{Model comparison across metrics}
        \label{fig:metrics_bar}
    \end{subfigure}
    \hfill
    \begin{subfigure}[t]{0.48\textwidth}
        \centering
        \includegraphics[width=\textwidth]{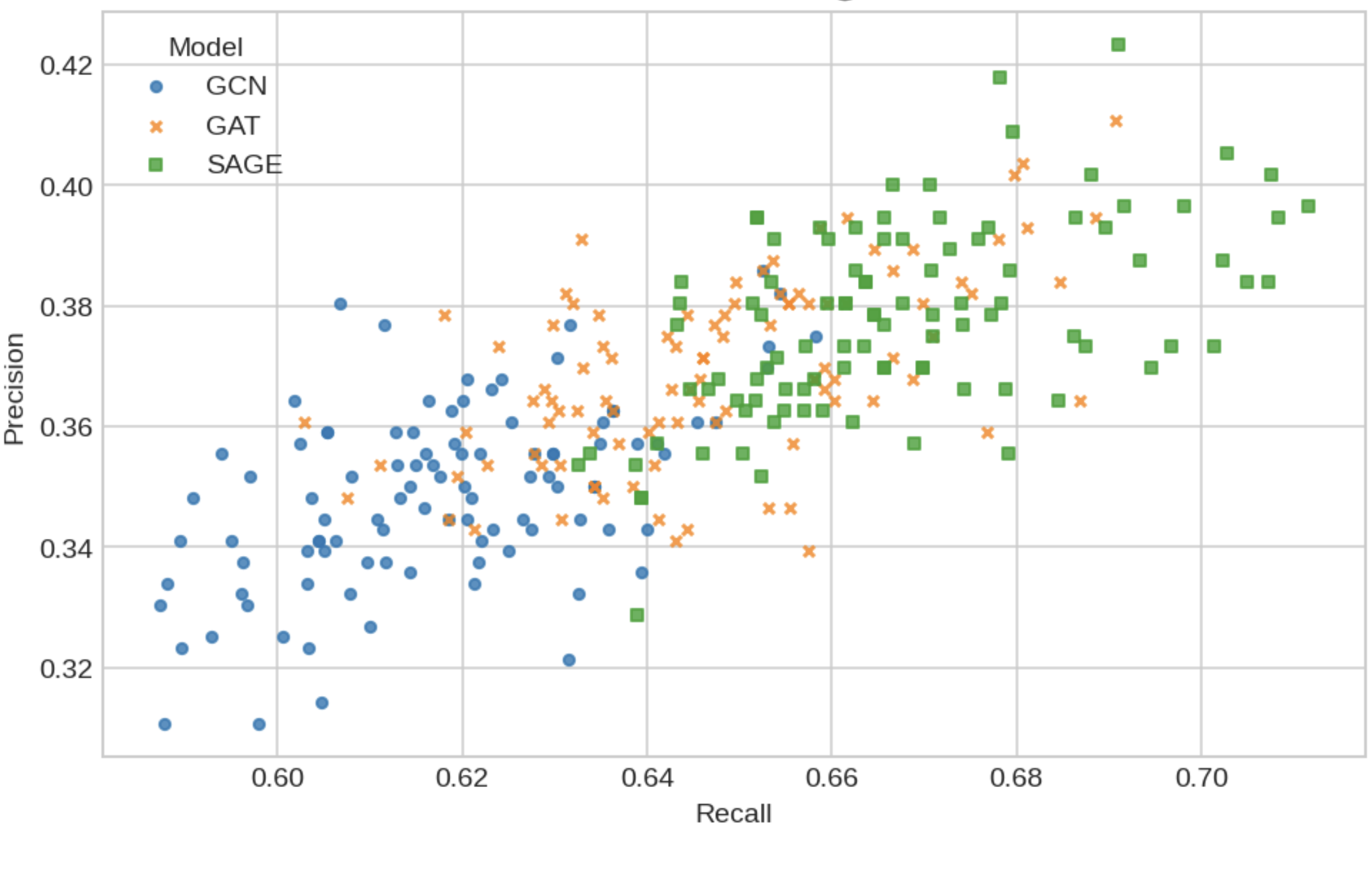}
        \caption{Precision--Recall trade-offs at 90\% confidence}
        \label{fig:pr_scatter}
    \end{subfigure}
    \caption{Performance comparison of GCN, GAT, and GraphSAGE on the Elliptic dataset. 
    (\subref{fig:metrics_bar}) Overall performance across metrics including AUC, AUPRC, and F1 at multiple thresholds. 
    (\subref{fig:pr_scatter}) Precision-Recall scatter plots at 90\% confidence, illustrating the distribution across architectures.}
    \label{fig:model_comparison}
\end{figure}

In addition to threshold-independent metrics, we also evaluate thresholded classification performance at high-confidence operating points (90th, 99th, and 99.9th percentiles of model output probabilities). At these thresholds, we compute precision, recall, and F1-score to quantify the trade-offs between coverage and reliability in fraud detection, as shown in Fig.~\ref{fig:model_comparison}.


\section{Results}\label{sec:results}
Table~\ref{tab:initialisation_norm} summarises the effect of initialisation and normalisation strategies across all three architectures. Results show that the optimal combination is architecture-specific. For GCN, the baseline configuration achieves the highest AUPRC (0.5993, AUC 0.8728); Xavier initialisation yields a marginal AUC improvement but reduces AUPRC, and adding GraphNorm depresses precision-recall performance further.

\begin{table}[t]
\centering
\caption{Effect of initialisation and GraphNorm on model performance. GCN achieves its best AUPRC at baseline (0.5993), with marginal changes under Xavier or GraphNorm. GAT benefits most from GraphNorm combined with Xavier ($+$0.055 AUPRC, $+$0.012 AUC over baseline). GraphSAGE reaches peak performance with Xavier initialisation alone ($+$0.013 AUPRC, $+$0.023 AUC over baseline), with GraphNorm providing diminishing returns.}
\label{tab:initialisation_norm}
\begin{tabular}{@{}lcccccc@{}}
\toprule
& \multicolumn{2}{c}{\textbf{Baseline}} & \multicolumn{2}{c}{\textbf{Xavier}} & \multicolumn{2}{c}{\textbf{GraphNorm + Xavier}} \\
\cmidrule(r){2-3} \cmidrule(lr){4-5} \cmidrule(l){6-7}
\textbf{Model} & AUC & AUPRC\phantom{88} & AUC & AUPRC\phantom{88} & \phantom{888}AUC & AUPRC \\
\cmidrule(r){1-1} \cmidrule(r){2-3} \cmidrule(lr){4-5} \cmidrule(l){6-7}
GCN       & \textbf{0.8728} & \textbf{0.5993} & 0.8740 & 0.5939 & \phantom{888}0.8736 & 0.5442 \\
GAT       & 0.8585 & 0.6022 & 0.8486 & 0.6190 & \phantom{888}\textbf{0.8700} & \textbf{0.6568} \\
GraphSAGE\phantom{88} & 0.8593 & 0.6551 & \textbf{0.8826} & \textbf{0.6678} & \phantom{888}0.8755 & 0.6651 \\
\bottomrule
\end{tabular}
\end{table}

GATs benefit substantially from combining GraphNorm with Xavier initialisation. The baseline GAT achieves AUC 0.8585 and AUPRC 0.6022. Xavier initialisation alone lowers AUC while modestly improving AUPRC. However, combining GraphNorm with Xavier yields the strongest GAT configuration, improving AUC to 0.8700 and AUPRC to 0.6568, suggesting that graph-level normalisation helps stabilise attention-based aggregation.

GraphSAGE responds most strongly to Xavier initialisation alone. Starting from a baseline of AUC 0.8593 and AUPRC 0.6551, Xavier initialisation yields the peak configuration at AUC 0.8826 and AUPRC 0.6678, gains of $+$0.023 AUC and $+$0.013 AUPRC. Adding GraphNorm on top of Xavier slightly reduces both metrics (AUC 0.8755, AUPRC 0.6651); in this case graph-level normalisation may interfere with GraphSAGE's neighbourhood aggregation.

\section{Discussion}\label{sec:discussion}
The results indicate that the structural characteristics of Elliptic-like transaction graphs, including heterophily, degree imbalance, and temporal drift, constrain the effectiveness of GNNs. These factors reduce neighbourhood feature consistency and hinder stable message propagation, explaining the limitations observed in model performance. This also addresses the first research question concerning properties that restrict GNN effectiveness on transaction data.

Initialisation strategies influence optimisation stability and convergence in an architecture-dependent manner. Xavier initialisation substantially improved GraphSAGE performance, yielding the largest gains across all evaluated configurations. For GCN, the default initialisation proved sufficient, with Xavier providing only marginal and inconsistent changes. These findings address the second research question regarding the role of initialisation in training stability.

Graph-specific normalisation demonstrated architecture-selective advantages. GraphNorm combined with Xavier initialisation produced the largest gains for GAT ($+$0.055 AUPRC over baseline), suggesting that graph-level normalisation helps stabilise attention-based aggregation under the skewed degree distribution of the Elliptic graph. However, GraphNorm did not benefit GCN or GraphSAGE; for the latter, it marginally reduced performance below the Xavier-only setting. This responds to the third research question on whether graph-specific normalisation better controls feature variation.

Across architectures, the optimal training strategy varies markedly. GraphNorm combined with Xavier initialisation benefits GAT most, Xavier initialisation alone is optimal for GraphSAGE, and GCN is best served by its default baseline configuration. These results address the fourth research question, indicating that no single normalisation or initialisation strategy dominates across all GNN architectures in the fraud detection context studied here.




\subsection{Comparison with Prior Work}
We situate our findings within the broader literature on GNN-based anti-money laundering using the Elliptic dataset. Prior studies have predominantly evaluated performance using threshold-dependent metrics such as F1-score. Weber et al.~\cite{weber2019amlgnn} reported F1 scores of 0.543 for logistic regression, 0.796 for random forest, and 0.628 for their GCN baseline. More recent work by Marasi and Ferretti~\cite{marasi2024} extended this comparison across multiple architectures, finding that ChebNet and GraphSAGE achieved F1 scores of 0.91 and 0.889 respectively, while GATv2 reached 0.881.\footnote{These F1 scores are computed over the full evaluation set at a default threshold of 0.5. On the Elliptic dataset, full-set F1 values are near 1 with variance $<10^{-5}$, making them uninformative for distinguishing model behaviour. We therefore follow the protocol of Deprez et al.~\cite{deprez2024aml} and report F1 at high-confidence thresholds (90th, 99th, 99.9th percentiles), as shown in Fig.~\ref{fig:model_comparison}.}

However, threshold-dependent metrics require selecting a specific classification threshold, introducing an additional hyperparameter that may not generalise across deployment contexts. In highly imbalanced settings such as the Elliptic dataset, where only 2\% of labeled transactions are illicit, threshold choice can substantially affect reported performance. A model that appears strong at one operating point may perform poorly at another.

We therefore prioritise AUPRC as our primary evaluation metric in Table~\ref{tab:sota_comparison}. Unlike F1-score, AUPRC summarises the precision--recall trade-off across all possible thresholds, providing a more complete picture of classifier performance under class imbalance~\cite{saito2015precision}. This is particularly relevant for AML applications, where practitioners must balance the cost of false positives (unnecessary investigations) against false negatives (missed illicit activity) according to operational constraints that vary across institutions.

\begin{table}[t]
\centering
\caption{Comparison with prior methods on the Elliptic dataset. Direct comparison is limited by differences in evaluation protocols.}
\label{tab:sota_comparison}
\begin{tabular}{lccc}
\toprule
\textbf{Method} & \textbf{AUC} & \textbf{AUPRC} & \textbf{Source} \\
\midrule
\multicolumn{4}{l}{\textit{Prior GNN Methods}} \\
GraphSAGE                       & 0.8712              & 0.6392            & Deprez et al.~\cite{deprez2024aml}\\
\midrule
\multicolumn{4}{l}{\textit{This Work}} \\
GAT (GraphNorm + Xavier)  \phantom{888}    & 0.8700              & 0.6568            & Ours \\
GCN (baseline)                      & 0.8728              & 0.5993            & Ours \\
GraphSAGE (Xavier)\phantom{888}     & \textbf{0.8826}     & \textbf{0.6678}   & Ours \\
\bottomrule
\end{tabular}
\end{table}

While direct numerical comparison with prior work is limited by these methodological differences, our results demonstrate that principled initialisation and normalisation can meaningfully improve GNN training dynamics in an architecture dependent manner. Our best-performing model, GraphSAGE with Xavier initialisation, achieves an AUPRC of 0.6678, surpassing the prior benchmark of 0.6392 reported by Deprez et al.~\cite{deprez2024aml}. GAT also benefits from GraphNorm combined with Xavier initialisation, reaching an AUPRC of 0.6568. Importantly, these improvements are orthogonal to architectural innovations explored in prior work, suggesting that combining principled training strategies with architectures such as ChebNet or GATv2 may yield additional benefits.

\section{Conclusion}
This study presents a systematic investigation into the role of weight initialisation and graph-specific normalisation in GNN-based anti-money laundering on the Elliptic dataset.

Our ablation of Xavier initialisation and GraphNorm across GCN, GAT, and GraphSAGE reveals that the effects of these training practices are architecture-dependent rather than universally beneficial. GraphSAGE achieves the strongest overall performance with Xavier initialisation alone (AUPRC 0.6678), GAT benefits most from combining GraphNorm with Xavier initialisation ($+$0.055 AUPRC over baseline), while GCN shows limited sensitivity to either modification, with baseline training achieving the highest precision--recall performance.

Overall, the findings suggest that training strategies such as initialisation and normalisation should be selected with respect to architectural inductive bias rather than assumed to yield uniform improvements. For imbalanced transaction graphs, careful initialisation and architecture-aware optimisation play a more consistent role than applying graph-level normalisation alone. These insights provide practical guidance for deploying GNNs in AML pipelines while maintaining stable training and reliable ranking performance.

\subsection{Limitations}
Although our study provides comprehensive insights into the effects of weight initialisation and graph-specific normalisation on GNN performance, several limitations persist in terms of both methodology and practical applicability.

We evaluated GCN, GAT, and GraphSAGE, which represent distinct inductive biases but do not cover the full range of modern designs. Graph transformers, higher-order GNNs, and temporal GNNs may respond differently to initialisation and normalisation. Moreover, the Elliptic dataset simplifies real financial networks; it lacks heterogeneous edge types, hierarchical account structures, and dynamic node attributes, so the observed gains may not fully transfer to operational AML systems.

Hardware limits constrained hyperparameter tuning to 100 Optuna trials within a bounded search space. We also evaluated only predictive metrics (AUC, AUPRC, precision, recall) without assessing inference latency, memory use, or throughput, which are important for real-time deployment at scale.

\subsection{Future Work}
Standardised evaluation protocols for the Elliptic benchmark may benefit the research community by adopting consistent temporal splits, metric definitions, and threshold selection procedures to enable more direct comparisons across studies.

Adaptive sampling methods such as GLASS~\cite{chen2021glass}, which uses reinforcement learning to identify informative subgraphs, could improve training efficiency but raise scalability concerns. Execution-level optimisations like SALIENT and SALIENT++ address some of these challenges, though SALIENT's partial C++ implementation complicates integration with Python-based ML pipelines. Developing lightweight, Python-compatible alternatives together with parallelism and distributed training would help build scalable GNN systems for financial anomaly detection.

Finally, evaluating our techniques on the more recent Elliptic2 dataset~\cite{bellei2024shape} could offer a more realistic and fine-grained view of money laundering at the subgraph level. It may reveal interactions between subgraph topology and training dynamics that are not visible in node-level datasets.

\begin{credits}
\subsubsection{\ackname} This work has been conducted through the RMIT Vietnam Research Grants-2025, administered by the Office for Research \& Innovation.

\subsubsection{\discintname}
The authors have no competing interests to declare that are
relevant to the content of this article.
\end{credits}
%
%
\bibliographystyle{splncs04}
\bibliography{references}
\end{document}